\documentclass[11pt]{article}
\usepackage[margin=1in]{geometry}
\usepackage{graphicx,booktabs,amsmath,amssymb,natbib,hyperref,xcolor}
\usepackage[T1]{fontenc}
\usepackage{tikz}
\usetikzlibrary{arrows.meta,positioning}
\hypersetup{colorlinks=true,linkcolor=blue!50!black,citecolor=blue!50!black,urlcolor=blue!50!black}
\title{Sequential knowledge editing breaks a model's ability to tell\\
good evidence from bad, without costing it accuracy}
\author{Atul Anand}
\date{}

\begin{document}
\maketitle

\begin{abstract}
Knowledge editing methods are evaluated on whether the edited fact changed, whether
paraphrases follow, and whether unrelated answers stayed put. We show that a model can pass
all three and still lose something the standard suite does not measure: the ability to
decide, on facts that were never edited, which retrieved documents to believe.

We score the log odds a model assigns to its own remembered answer against the answer
asserted by an injected passage, before and after editing, holding the query, the passage,
and both candidate strings fixed. Our cleanest arm is a conservatively tuned LoRA: after
1{,}000 sequential edits on Qwen2.5-7B-Instruct it leaves MMLU unchanged to four decimal
places, and the spread of the arbitration quantity across untouched facts still falls by
36\%. MEMIT shows the same pattern at a small capability cost (MMLU 0.6275 to 0.6050)
rather than none. Selective prediction degrades accordingly:
the area under the risk-coverage curve rises by 0.107, against 0.005 for a
norm-matched random perturbation at the same MMLU. Error on the model's most confident
quarter of arbitration decisions goes from 0.217 to 0.342 while overall error rises far
less, so accuracy understates the harm roughly threefold.

The effect is not capability loss. We sweep random perturbation over five severities and
find that damage bad enough to cut MMLU from 0.6275 to 0.3725 produces less harm
(0.088) than MEMIT does at 0.6050 (0.102). It survives across three seeds, two model
families, two datasets, two probe-disjointness criteria, three prompt templates, and
paraphrased queries. Layer ablation on MEMIT's saved weight deltas shows the effect is
distributed: no single layer reproduces it, and removing any one recovers about half.
In end-to-end retrieval with a frozen retriever, accuracy falls from 0.592 to 0.46 and
the controlled measurement ranks checkpoints correctly.

A secondary finding may matter more to practitioners. Three of five model and method
pairings we ran collapse to chance MMLU at 1{,}000 sequential edits using their
published hyperparameters, while edit success stays at 1.00 and locality looks clean.
Sequential-editing evaluations that never measure capability cannot see this.
\end{abstract}

\section{Introduction}

An edited model does not answer from its weights alone. In deployment it answers from its
weights and whatever a retriever hands it, and it has to decide which to trust when the two
disagree. That decision is a policy, and nothing in the standard editing evaluation asks
whether editing changes it.

The question is not whether an edited fact survives contradicting context. That has been
studied. The question here is whether editing a thousand facts changes how the model
arbitrates on the other facts, the ones nobody touched. If it does, the damage is invisible
to edit success, to paraphrase generalization, and to neighborhood locality, because all
three ask about answers rather than about how answers get chosen.

We find that it does, and that the change is not explained by the model getting worse.
Our clearest arm finishes 1{,}000 sequential edits with MMLU identical to the unedited
baseline and still shows a twentyfold increase in selective-prediction error relative to a
matched random perturbation. A model that is exactly as good at multiple-choice reasoning
has become much worse at knowing when to believe a document.

\paragraph{Contributions.}
We define an arbitration measurement that is paired at the level of the individual fact and
context, and show that its mean is the wrong summary: the effect is a collapse in spread,
not a shift in center, and the mean cancels. We show the collapse is editing-specific
against a damage-response curve rather than a single matched control. We give a causal
localization by adding and removing exact weight deltas. We connect the controlled
measurement to end-to-end retrieval failure. Finally we report that several published
editor configurations destroy the model at edit loads the literature treats as routine.

\section{Related work}

Work on context-robust editing asks whether an edited fact survives contradicting context
\citep{contextrobust2025}. We ask the reverse: whether editing changes context reliance on
facts that were never edited.

A separate line reports that editing leaves models underconfident on the facts it edited,
diagnosed through the gap between token probability and accuracy \citep{underconfidence2025}.
That is calibration on edited facts. Ours is arbitration between memory and an external
document on facts left alone, which is a different quantity on a different population, and
our no-edit and damage-matched controls are aimed at a different confound. The two results
are compatible and we do not read either as subsuming the other. Work on sequential editing has documented noise accumulation
and spectral collapse \citep{noise2025,spectral2026}. Those results concern the model's own
knowledge; ours concerns its treatment of external evidence, and our degradation controls
show the two come apart. AlphaEdit adds null-space constraints intended to preserve
unrelated knowledge \citep{alphaedit2024}; we find its standard locality behavior does not
imply stable arbitration, and that at 1{,}000 sequential edits on Qwen2.5-7B it falls to
chance MMLU.

% Method diagram, TikZ so it stays vector and rebuilds from source.
%
% Topology matters more than styling here. An earlier version put editor, conditions and
% scoring in one column, so the arrow from editor to scoring ran straight through the
% conditions box. This version uses two parallel lanes that converge once, which makes it
% impossible for an arrow to cross a node.
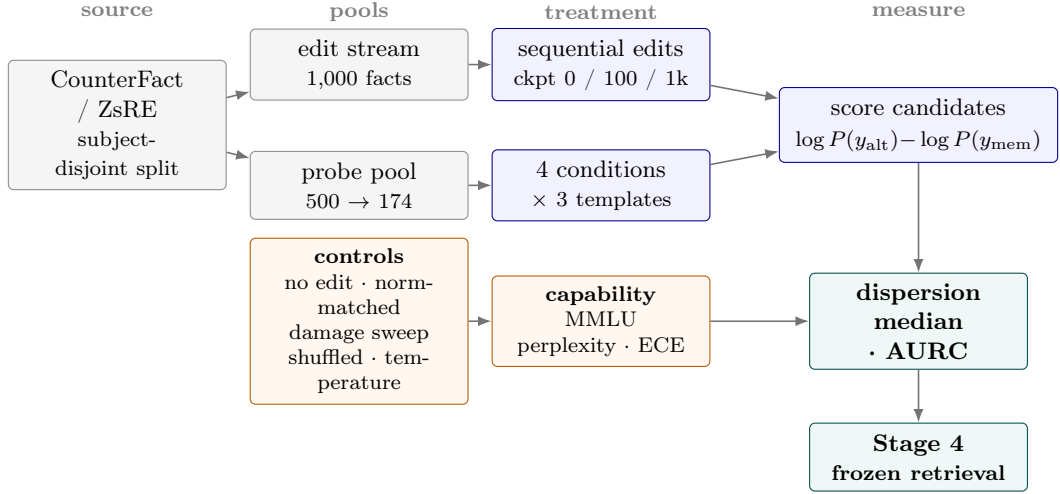
\begin{figure}[t]
\centering
\begin{tikzpicture}[
  font=\footnotesize,
  box/.style  ={draw=black!45, rounded corners=2pt, align=center, inner sep=4pt,
                minimum height=9mm, text width=26mm, fill=white},
  data/.style ={box, fill=black!4},
  proc/.style ={box, draw=blue!55!black, fill=blue!5},
  ctrl/.style ={box, draw=orange!70!black, fill=orange!6, font=\scriptsize},
  res/.style  ={box, draw=teal!60!black, fill=teal!6, font=\footnotesize\bfseries},
  ar/.style   ={-{Latex[length=1.8mm]}, draw=black!55, semithick},
  hdr/.style  ={font=\scriptsize\bfseries, text=black!50},
]
% four columns, two lanes, one convergence
\def\cA{0} \def\cB{32} \def\cC{64} \def\cD{106}
\def\up{0} \def\dn{-16} \def\mid{-8}

\node[hdr] at (\cA mm,  7mm) {source};
\node[hdr] at (\cB mm,  7mm) {pools};
\node[hdr] at (\cC mm,  7mm) {treatment};
\node[hdr] at (\cD mm,  7mm) {measure};

% single source, forking into two lanes
\node[data, anchor=center] (cf) at (\cA mm, \mid mm)
      {CounterFact / ZsRE\\\scriptsize subject-disjoint split};

% upper lane: the edits
\node[data, anchor=center] (edits)  at (\cB mm, \up mm) {edit stream\\\scriptsize 1{,}000 facts};
\node[proc, anchor=center] (editor) at (\cC mm, \up mm) {sequential edits\\\scriptsize ckpt 0 / 100 / 1k};

% lower lane: the probes
\node[data, anchor=center] (probes) at (\cB mm, \dn mm) {probe pool\\\scriptsize 500 $\to$ 174};
\node[proc, anchor=center] (cond)   at (\cC mm, \dn mm) {4 conditions\\\scriptsize $\times$ 3 templates};

% the lanes converge exactly once
\node[proc, anchor=center, text width=34mm] (score) at (\cD mm, \mid mm)
      {score candidates\\[1pt]\scriptsize $\log P(y_{\mathrm{alt}}){-}\log P(y_{\mathrm{mem}})$};

% results, straight down the right column
\node[res, anchor=center] (metrics) at (\cD mm, -34mm) {dispersion\\median · AURC};
\node[res, anchor=center] (rag)     at (\cD mm, -52mm) {Stage 4\\\scriptsize frozen retrieval};

% controls, on their own row, feeding the results from the left
\node[ctrl, anchor=center] (c1) at (\cB mm, -34mm)
      {\textbf{controls}\\ no edit · norm-matched\\ damage sweep\\ shuffled · temperature};
\node[ctrl, anchor=center] (c2) at (\cC mm, -34mm)
      {\textbf{capability}\\ MMLU\\ perplexity · ECE};

\draw[ar] (cf) -- (edits);
\draw[ar] (cf) -- (probes);
\draw[ar] (edits)  -- (editor);
\draw[ar] (probes) -- (cond);
\draw[ar] (editor) -- (score);
\draw[ar] (cond)   -- (score);
\draw[ar] (score)   -- (metrics);
\draw[ar] (metrics) -- (rag);
\draw[ar] (c1) -- (c2);
\draw[ar] (c2) -- (metrics);
\end{tikzpicture}
\caption{Experimental design. The edit stream and the probe pool share no subject entity,
so everything measured downstream concerns facts the editor never touched. Controls and the
capability battery run at every checkpoint rather than once at the end. Retrieval in
Stage~4 is computed once and replayed for each checkpoint, which is what makes the accuracy
differences attributable to the model rather than to what was retrieved.}
\label{fig:method}
\end{figure}

\section{Measuring arbitration}

Fix an untouched factual query $q$. Let $y_{\text{mem}}$ be the answer the unedited model
gives with no context, and $y_{\text{alt}}$ a competing object. For a context $c$ we score
both candidate strings under the same prefix and record
\[
M(q,c) \;=\; \log P_\theta(y_{\text{alt}} \mid q,c) \;-\; \log P_\theta(y_{\text{mem}} \mid q,c).
\]
$M$ does not depend on which candidate the passage happens to assert, which lets one
primitive serve every evidence condition. In analysis we orient it toward the asserted
object to obtain the arbitration term $A$, and the paired drift is
$A_{\text{after}} - A_{\text{before}}$ with $q$, $c$, and both strings held fixed.

This is a forced two-candidate contrast rather than open-ended generation. Scoring free
text would confound arbitration with formatting,
verbosity, and refusal behavior, all of which editing also perturbs; the constrained
contrast isolates the decision we care about, and it is the same construction the editing
literature already uses to define edit success. The obvious objection is that a constrained
probe need not predict unconstrained behavior. We do not assume it does. Section~7 tests it
directly against open-ended answers under real retrieval, and the ordering holds across
five checkpoints and two editors. Readers who reject the probe on principle can read
Section~7 on its own, since its accuracy numbers involve no log-odds scoring at all.

\subsection{Four evidence conditions, never collapsed}

For each probe we build a passage that agrees with memory, one that asserts the alternative
as a genuine update, one that asserts it as a plain (and false) claim, and one that asserts
it with unsupported authority. The agreeing and stale-false passages use an identical
wrapper and differ only in the asserted object, which separates "trusts the context" from
"prefers this particular string". The three conflicting passages assert the same object and
differ only in framing.

Collapsing these into one override rate would hide the result. A model that grows more
credulous improves on true updates and degrades under adversarial text at the same time.

\subsection{Why the mean is the wrong statistic}

In our pilot the mean paired drift on stale-false evidence at 1{,}000 LoRA edits was
$-2.70$, the median was $-6.06$, and the standard deviation of $A$ fell from 12.08 to 5.56.
The distribution is heavy-tailed, so facts crossing up from the negative tail cancel facts
moving down and the mean understates the change. The override rate rose over the same
interval, which reads as more context trust while the model was in fact losing
fact-specific discrimination.

We therefore report the dispersion ratio $sd_{\text{after}}/sd_{\text{before}}$ and, as the
headline, the area under the risk-coverage curve.

\subsection{Selective prediction}

Dispersion is not self-evidently harmful, so we cash it out. A deployed system can abstain.
That only works if the arbitration margin orders decisions, with large $|A|$ where the model
is right. We sort by $|A|$, sweep coverage, and record risk among answered queries; AURC is
the area under that curve.

AURC is not a restatement of dispersion. Rescaling the margin preserves ordering and leaves
AURC unchanged to machine precision, which our test suite asserts directly. It moves only
when editing reorders which facts the model is confident about, so a compression that kept
the ranking intact would score zero here.

\section{Setup}

Qwen2.5-7B-Instruct is the primary model, with Mistral-7B-Instruct-v0.3 for replication.
Edits come from CounterFact, with ZsRE as a second dataset. The probe pool shares no subject
with the edit stream and is relation- and answer-length stratified against it; we also run a
split sharing no relation either. Of 500 probes, 174 pass the primary filter on Qwen (the
base model must be confidently correct under all three prompt templates), and 167 on
Mistral.

Editors are LoRA at two settings, and MEMIT and AlphaEdit through EasyEdit at commit
\texttt{14cea82}, using upstream hyperparameters unchanged. We reduced the second-moment
sample count from 100{,}000 to 20{,}000 for tractability; both locate-then-edit methods read
the same cached statistics, so the contrast between them is unaffected. LoRA runs one edit
at a time and MEMIT in closed-form batches of 100. The asymmetry is deliberate: each editor
runs in its own natural regime rather than a shared one that handicaps one of them.

Seeds vary the order of the same 1{,}000 edits. This matters because MEMIT and AlphaEdit are
deterministic closed-form solves, so without reordering they return identical results at
every seed and appear perfectly reproducible while measuring nothing.

\subsection{Controls}

\begin{itemize}
\item \textbf{No edit.} Identical evaluation with no update. Reproduces the baseline to
      machine precision, which is the check that the pipeline cannot manufacture drift.
\item \textbf{Norm-matched perturbation.} Random low-rank noise in the same layers at the
      realized per-layer update norms of the arm it matches.
\item \textbf{Damage-response sweep.} The same noise at five severities, giving a curve
      rather than one comparison point.
\item \textbf{Shuffled targets.} The identical editing procedure with each edit trained
      toward another fact's target under a seeded derangement.
\item \textbf{Temperature-matched base model.} A single scalar fitted so the unedited model
      matches the edited model's confidence.
\end{itemize}

Two of these produced negative results we report rather than discard. The shuffled-target
control reproduced the entire dispersion collapse of the aggressive LoRA arm at 100 edits
(0.45 against 0.49), which is why that arm is reported as a fine-tuning artifact and moved
out of the main results. At 1{,}000 edits the same control stops being informative, because
inserting incoherent bindings through a closed-form solve destroys the model.

We also note that a norm-matched control only controls for anything if the arm it matches is
sane. On Mistral, the aggressive LoRA configuration explodes, and the perturbation matched
to it is destroyed along with it.

\section{Results}

\begin{figure}[t]
\centering
\includegraphics[width=0.72\textwidth]{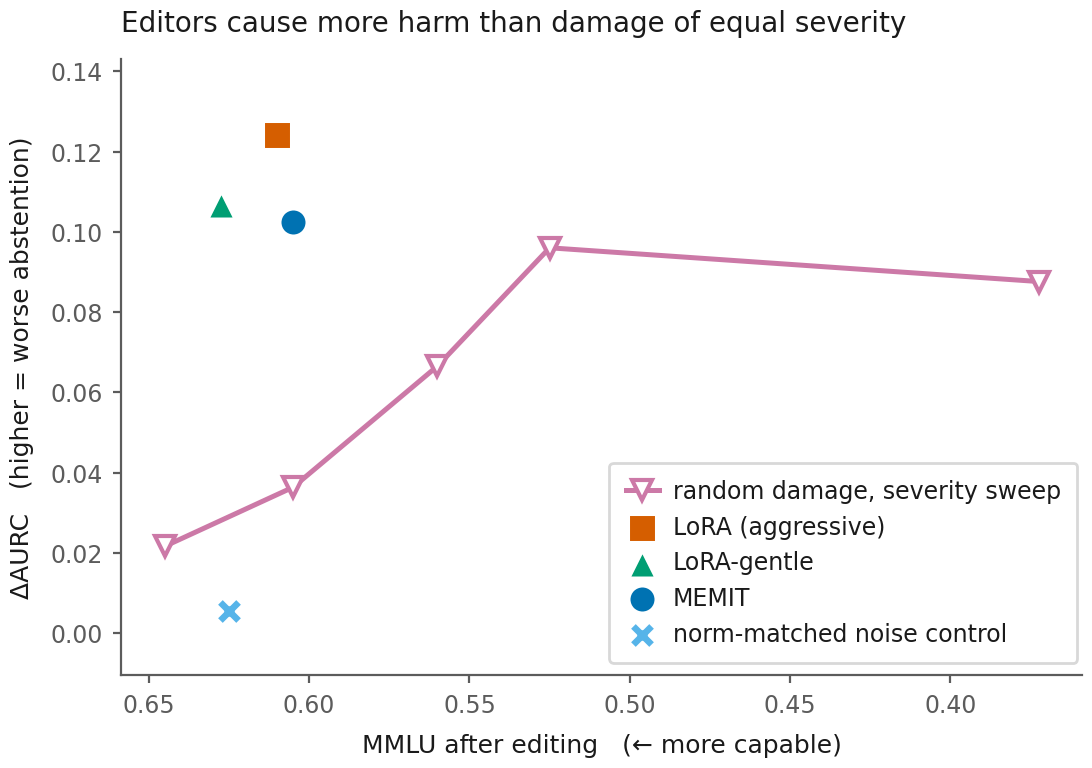}
\caption{$\Delta$AURC against MMLU after editing. The curve traces what random perturbation
alone produces as its severity increases. All three editors lie well above it at comparable
or better capability, and the norm-matched control lies on the floor. Damage severe enough
to cut MMLU to 0.3725 produces less arbitration harm than MEMIT does at 0.6050.}
\label{fig:damage}
\end{figure}

\begin{table}[t]
\centering
\small
\begin{tabular}{lrrrr}
\toprule
Arm (1{,}000 edits, Qwen) & MMLU & PPL & $sd_{\text{after}}/sd_{\text{before}}$ & $\Delta$AURC \\
\midrule
No edit                & 0.6275 &  9.80 & 1.000 & $+0.0000$ \\
Norm-matched noise     & 0.6250 &  9.90 & 0.978 & $+0.0054$ \\
LoRA-gentle            & 0.6275 & 28.96 & 0.640 & $+0.1067$ \\
MEMIT                  & 0.6050 & 11.49 & 0.622 & $+0.1024$ \\
LoRA (aggressive)      & 0.6100 & 39.23 & 0.460 & $+0.1241$ \\
AlphaEdit              & 0.2750 & 30.70 & 0.534 & $+0.1636$ \\
\bottomrule
\end{tabular}
\caption{Dispersion is on stale-false evidence. LoRA-gentle reaches the unedited MMLU value
exactly and still causes twenty times the selective-prediction harm of a perturbation at the
same MMLU. AlphaEdit's MMLU of 0.2750 is chance for four-way multiple choice, so its
$\Delta$AURC is not an arbitration result.}
\label{tab:main}
\end{table}

Table~\ref{tab:main} carries the central claim. LoRA-gentle, whose hyperparameters were
chosen on edit success and locality alone, finishes 1{,}000 sequential edits at MMLU 0.6275,
which is the unedited value to four decimals, and still shows $\Delta$AURC of $+0.1067$
against $+0.0054$ for the matched control.

Perplexity does rise, from 9.80 to 28.96, so the arm is not literally damage-free. We could
not find a damage axis that accounts for the gap, though. The sweep reaches comparable
perplexity at its most severe setting (25.97), and there it produces less harm ($+0.0876$)
while MMLU collapses to 0.3725.

AURC does not care about direction. LoRA-gentle drifts $-8.47$ in median terms (more
stubborn) and MEMIT drifts $+1.69$ (more credulous), yet both land near $+0.10$. Whatever
is being lost, it is not a preference for one answer over the other; it is the model's
ability to tell the two situations apart.

\subsection{Dispersion separates from damage with intervals}

Comparing MEMIT to a damage-matched perturbation (MMLU 0.565 against MEMIT's 0.605, so
slightly worse), the difference in dispersion ratio is $-0.366$ with a 95\% cluster-bootstrap
interval of $[-0.452,-0.276]$ on stale-false evidence, $-0.259\,[-0.363,-0.151]$ on true
updates, and $-0.247\,[-0.328,-0.160]$ on adversarial text. On agreeing evidence the
interval contains zero.

Notably the damage-matched control reproduces most of MEMIT's median drift, at $+1.25$
against $+1.69$. Only the dispersion channel separates them.

\begin{figure}[t]
\centering
\includegraphics[width=0.68\textwidth]{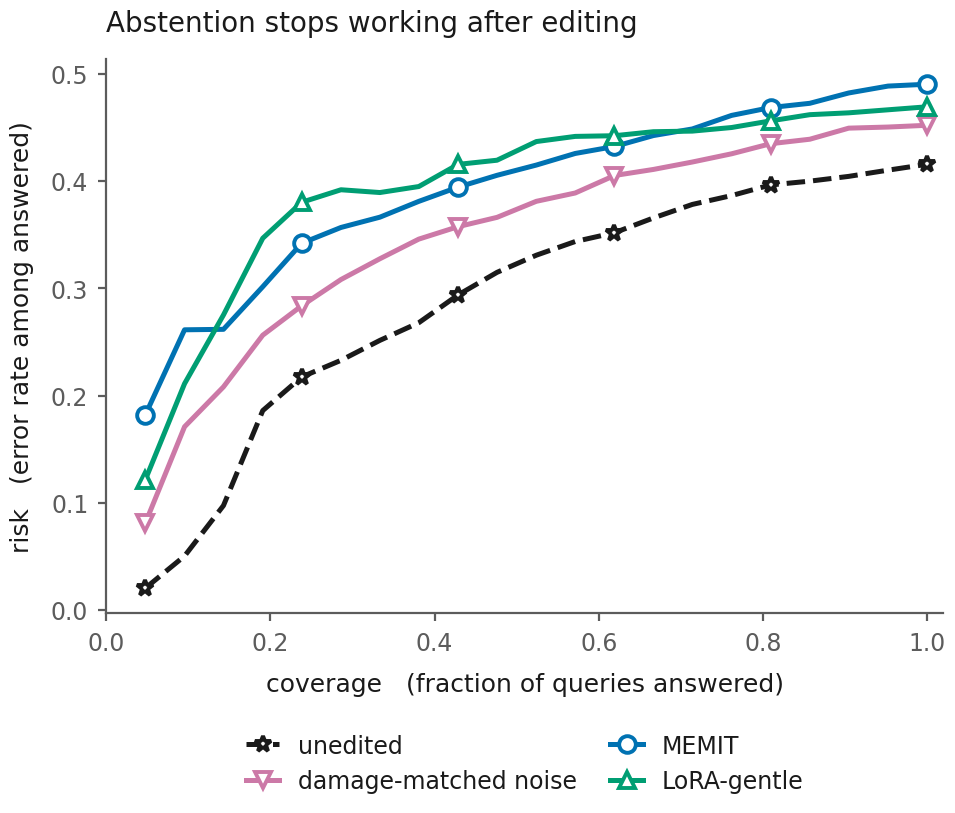}
\caption{Risk-coverage curves. The unedited model (dashed) keeps error low when it answers only its most confident queries. After editing that ordering degrades, and the gap is widest exactly where a deployed system would operate, at low coverage.}
\label{fig:rc}
\end{figure}

\begin{figure}[t]
\centering
\includegraphics[width=0.75\textwidth]{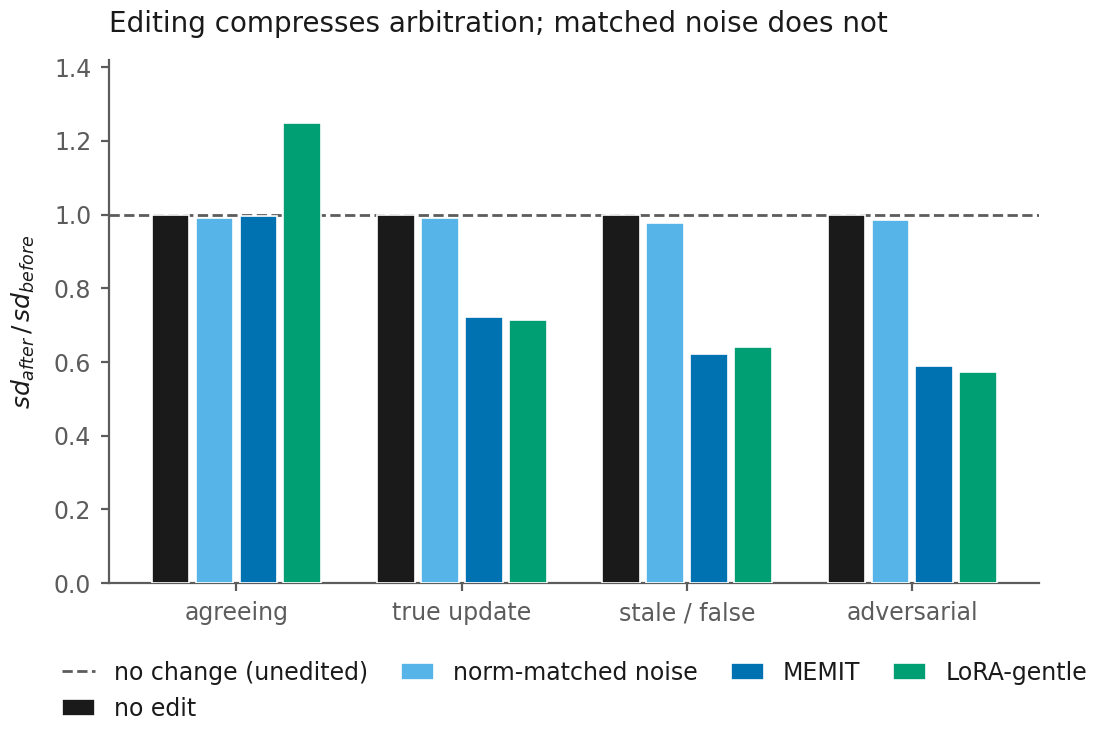}
\caption{Dispersion ratio by evidence condition. Matched noise sits at 1.0 in every condition. Both editors compress the three conflicting conditions to roughly 0.6. The agreeing condition, where memory and passage point the same way, behaves differently and is the one condition where the editor-versus-noise interval contains zero.}
\label{fig:disp}
\end{figure}

\subsection{A joint model over editors and conditions}

Bootstrap intervals treat each arm on its own. Because the same 174 facts appear in every
arm, the residuals are correlated in a way per-arm intervals cannot express, so we also fit
a mixed model with a random intercept per fact (Table~\ref{tab:mixed}).

Every editor separates from the no-edit reference except the norm-matched control, whose
coefficient is $+0.14$ with $p=0.60$. The editor-by-condition interactions are large and
ordered the way the per-arm analysis suggested. Seed is a fixed effect rather than random:
three levels cannot support a stable variance component, and we would rather say that
plainly than report a number the design does not license.

\begin{table}[t]
\centering\small
\begin{tabular}{lrrrr}
\toprule
Term & Coef. & SE & $z$ & $p$ \\
\midrule
AlphaEdit & -8.73 & 0.27 & -32.2 & $<$0.001 \\
LoRA & -15.55 & 0.27 & -57.4 & $<$0.001 \\
LoRA-gentle & -11.31 & 0.33 & -34.7 & $<$0.001 \\
MEMIT & -2.31 & 0.27 & -8.5 & $<$0.001 \\
norm-matched & +0.14 & 0.27 & +0.5 & 0.604 \\
load 1000 & -2.25 & 0.06 & -36.4 & $<$0.001 \\
seed 1 & -1.62 & 0.08 & -19.8 & $<$0.001 \\
seed 2 & +1.10 & 0.08 & +13.5 & $<$0.001 \\
AlphaEdit $\times$ adversarial & +7.08 & 0.38 & +18.8 & $<$0.001 \\
LoRA $\times$ adversarial & +11.54 & 0.38 & +30.6 & $<$0.001 \\
LoRA-gentle $\times$ adversarial & +7.71 & 0.46 & +16.7 & $<$0.001 \\
MEMIT $\times$ adversarial & +5.90 & 0.38 & +15.7 & $<$0.001 \\
norm-matched $\times$ adversarial & +0.21 & 0.38 & +0.6 & 0.579 \\
AlphaEdit $\times$ stale & +6.76 & 0.38 & +17.9 & $<$0.001 \\
LoRA $\times$ stale & +10.55 & 0.38 & +28.0 & $<$0.001 \\
LoRA-gentle $\times$ stale & +5.59 & 0.46 & +12.1 & $<$0.001 \\
MEMIT $\times$ stale & +5.39 & 0.38 & +14.3 & $<$0.001 \\
norm-matched $\times$ stale & +0.42 & 0.38 & +1.1 & 0.264 \\
AlphaEdit $\times$ true upd. & +5.01 & 0.38 & +13.3 & $<$0.001 \\
LoRA $\times$ true upd. & +7.57 & 0.38 & +20.1 & $<$0.001 \\
LoRA-gentle $\times$ true upd. & +2.50 & 0.46 & +5.4 & $<$0.001 \\
MEMIT $\times$ true upd. & +4.07 & 0.38 & +10.8 & $<$0.001 \\
norm-matched $\times$ true upd. & +0.27 & 0.38 & +0.7 & 0.467 \\
\bottomrule
\end{tabular}
\caption{Mixed-effects fit on paired drift, with a random intercept per fact. 58{,}464 observations over 174 facts; the model converged. Reference levels are the no-edit arm and the agreeing condition, so each editor coefficient reads as a departure from applying no update at all. The norm-matched control is the only editor term whose interval includes zero. Seed enters as a fixed effect rather than random: three levels are too few to estimate a variance component stably.}
\label{tab:mixed}
\end{table}

\subsection{Generality}

\begin{table}[t]
\centering
\small
\begin{tabular}{llrrr}
\toprule
Axis & Setting & $n$ & LoRA-gentle $\Delta$AURC & Control $\Delta$AURC \\
\midrule
Primary            & CounterFact, subject-disjoint       & 174 & $+0.1067$ & $+0.0054$ \\
Probe split        & subject and relation disjoint        & 196 & $+0.1182$ & $+0.0016$ \\
Query phrasing     & paraphrased stems                    & 108 & $+0.1348$ & $+0.0009$ \\
Dataset            & ZsRE                                 &  66 & $+0.0743$ & $+0.0010$ \\
Model family       & Mistral-7B-Instruct-v0.3             & 167 & $+0.0897$ & $+0.0039$ \\
\bottomrule
\end{tabular}
\caption{The effect holds on every axis we varied, against a control that stays near zero
throughout. Sample sizes differ because each split applies its own confidence filter.}
\label{tab:general}
\end{table}

\begin{figure}[t]
\centering
\includegraphics[width=0.60\textwidth]{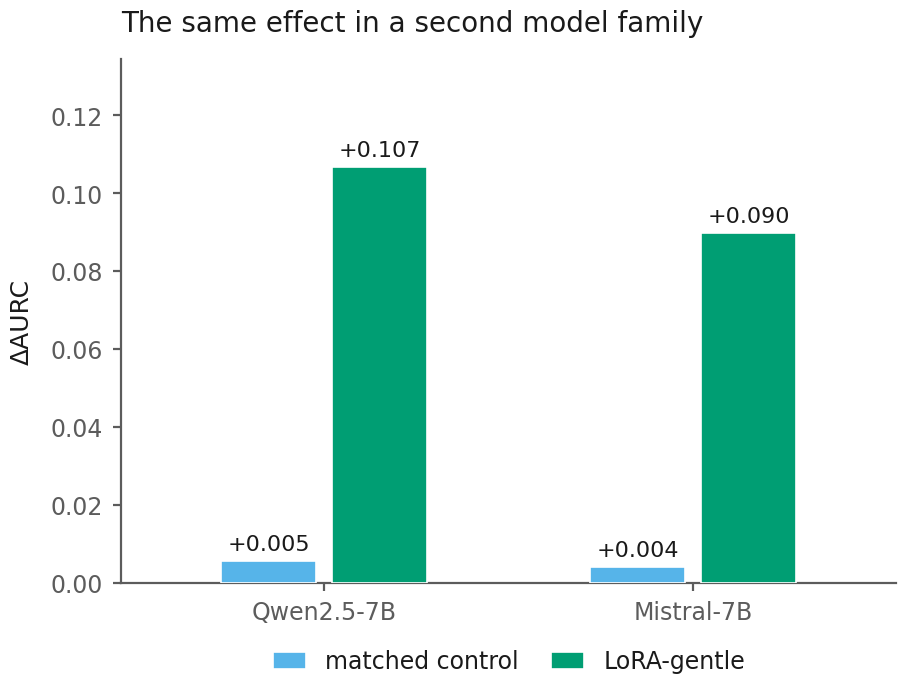}
\caption{The same comparison on both model families. Absolute magnitude differs; the separation from the matched control does not.}
\label{fig:cross}
\end{figure}

Seeds vary the edit order. On Qwen the effect is stable: MEMIT gives dispersion 0.622 and
0.628 across two seeds with $\Delta$AURC $+0.1024$ and $+0.1072$; the aggressive LoRA arm
gives $+0.1241$, $+0.1166$, and $+0.1043$ across three.

On Mistral it is not. LoRA-gentle gives $+0.0897$, $+0.1493$, and $+0.0467$ across three
seeds, a spread of more than threefold that differs only in the order the same 1{,}000
edits arrive. We take this as showing existence and direction on a second family, not a
magnitude. Every seed clears the largest control seed by more than tenfold, so the
separation is not in question, but three seeds cannot pin the size of an effect that varies
this much and we do not report a Mistral mean. Establishing the magnitude there needs more
seeds than we ran.

AURC is the most seed-stable of the three statistics we tried. The median swung from
$-6.06$ to $-12.09$ across seeds for the same arm.

\section{The effect is distributed across layers}

MEMIT writes to five \texttt{mlp.down\_proj} matrices. We save those weight deltas, reload
the base model, and re-apply subsets, so nothing is replayed and nothing is correlational.

Applying any single layer's update alone gives a dispersion ratio between 0.97 and 1.03
against 0.62 for the full update, so one layer on its own does essentially nothing.
Removing any single layer recovers only about half the effect, leaving 0.79 to 0.85.
Summed individually the five contributions come to roughly 0.03 of compression; together
they produce 0.38.

So the update has to be read jointly. That is consistent with the norm-matched
perturbation being inert, and it means a layer-by-layer analysis would have found nothing
here.

\begin{figure}[t]
\centering
\includegraphics[width=0.80\textwidth]{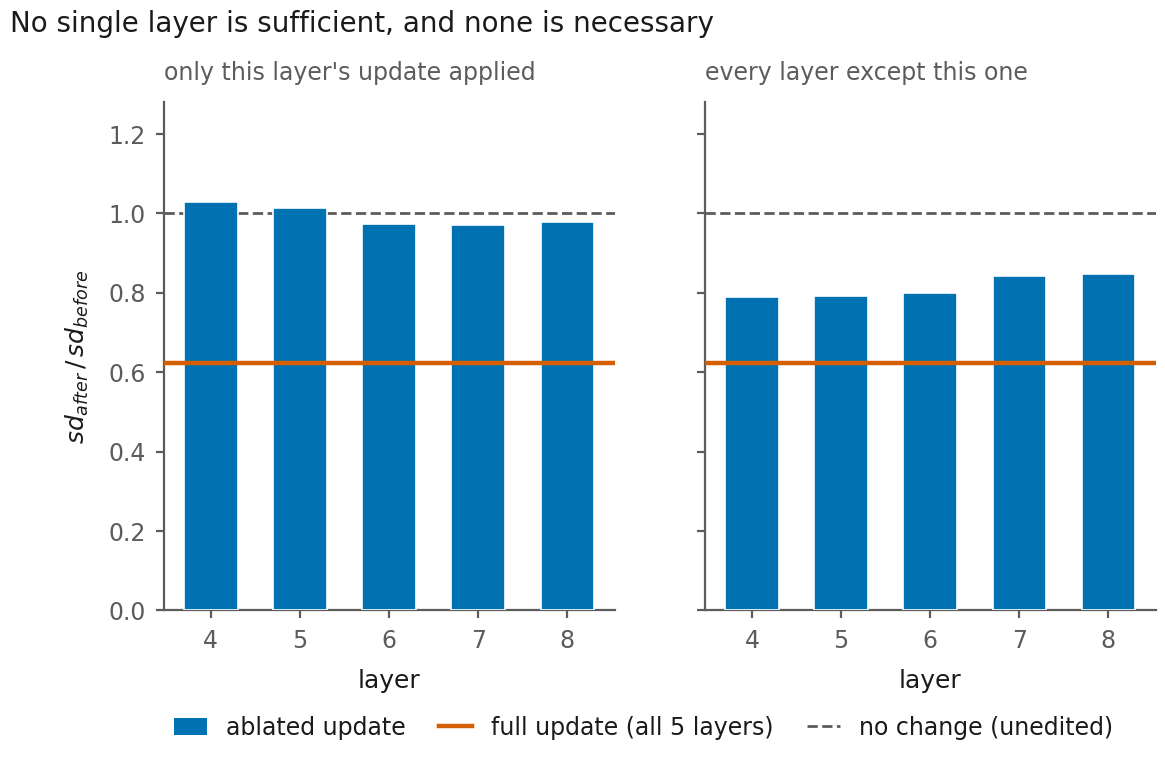}
\caption{Layer ablation on MEMIT's saved deltas. Left: only one layer's update applied. Right: every layer except one. Neither panel approaches the full-update line, so the effect is carried by the joint update rather than by any individual layer.}
\label{fig:abl}
\end{figure}

\section{End-to-end retrieval}

We index a corpus in which each probe has a correct document, a stale one, and an
adversarial one, plus 600 distractors, and retrieve top-3 once with a frozen BGE-M3 and
FAISS. The same passage set is replayed for every checkpoint, so nothing below is
attributable to retrieval quality. Every query retrieves all three of its competing
documents, which makes this a stress test rather than a naturalistic distribution.

Accuracy falls from 0.592 unedited to 0.454 for MEMIT and 0.471 for LoRA-gentle at 1{,}000
edits, with the model following a misleading document on the majority of queries. AlphaEdit
reaches 0.029. Across five checkpoints spanning two editors, Stage-1 $\Delta$AURC orders
end-to-end accuracy correctly.

\begin{figure}[t]
\centering
\includegraphics[width=0.68\textwidth]{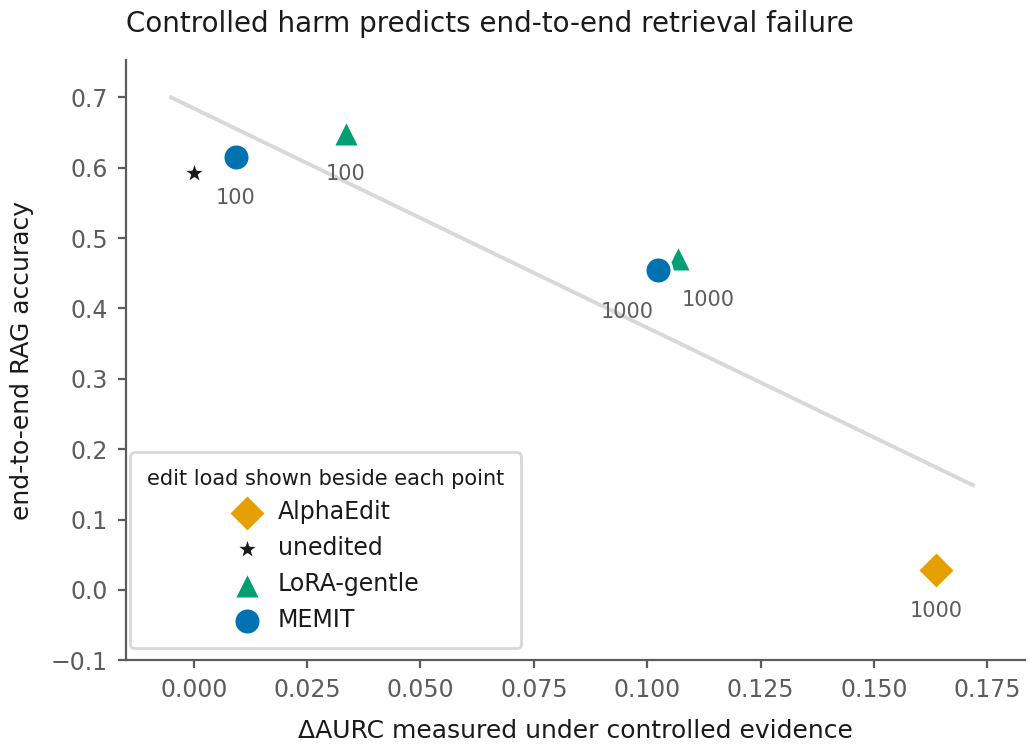}
\caption{Controlled arbitration harm against end-to-end retrieval accuracy, with retrieval held fixed across checkpoints. The number beside each point is the edit load.}
\label{fig:s4}
\end{figure}

Per fact, the picture is mixed and we state it plainly. For LoRA-gentle the post-edit
arbitration margin predicts which specific facts fail, with AUC 0.701 against 0.523 for
pre-edit confidence. For MEMIT it does not, at 0.513 against 0.603. Controlled arbitration
measurements indicate which models will fail end-to-end, not reliably which facts.

\section{Published configurations that destroy the model}

Three of the five model and method pairings we ran fall to chance MMLU at 1{,}000
sequential edits, using upstream hyperparameters: AlphaEdit on Qwen (0.2750), MEMIT on
Mistral (0.2425), and aggressive LoRA on Mistral (0.2125). MEMIT survives on Qwen at 0.6050,
so which pairing holds up looks idiosyncratic rather than principled.

Throughout these runs edit success stayed at 1.00 and neighborhood locality looked
unremarkable. A sequential-editing evaluation that does not measure capability cannot
distinguish a working editor from a destroyed model.

\section{Limitations}

Per-fact prediction of retrieval failure works for the LoRA arm and not for MEMIT, so the
deployment claim is about models rather than individual facts. Magnitude on Mistral varies
threefold across seeds. Stage 4 is maximally conflicted by construction. The ZsRE panel has
66 probes after filtering, small enough that it should be read as agreement in direction
rather than a second independent estimate. We reduced MEMIT's second-moment sample count
below the published setting. AlphaEdit ships no Mistral configuration, so the replication
panel is LoRA and MEMIT only. Finally, our conclusion that the aggressive LoRA arm is an
artifact rests on a control that stops working at the higher edit load.

\section{Conclusion}

After a thousand edits our conservatively tuned LoRA arm answers multiple-choice questions
exactly as well as it did before, and is substantially worse at judging which retrieved
document to believe. The standard editing evaluation does not detect this, because it asks
what the model answers rather than how it chooses.

For a practitioner this is a model-level warning rather than a per-answer one. The
controlled measurement ranks checkpoints by how badly their retrieval behavior will
degrade. It does not reliably say which individual answer to distrust, and on MEMIT it does
not beat pre-edit confidence at that task at all.

\appendix
\section{Capability at every checkpoint}

Table~\ref{tab:degradation} lists MMLU, expected calibration error and perplexity for all
57 Qwen checkpoints. We include it in full because the central claim depends on capability
being held constant, and a reader should be able to check that rather than take the four
rows quoted in the main text on trust.

\begin{table}[t]
\centering\footnotesize
\begin{tabular}{lrrr}
\toprule
Checkpoint & MMLU & ECE & PPL \\
\midrule
alphaedit L0 & 0.6275 & 0.2962 & 9.80 \\
alphaedit L0 (s1) & 0.6275 & 0.2962 & 9.80 \\
alphaedit L1000 & 0.2750 & 0.2734 & 30.70 \\
alphaedit L1000 (s1) & 0.2450 & 0.4104 & 1766.24 \\
alphaedit L1000 (s2) & 0.5175 & 0.4035 & 15.31 \\
alphaedit L100 & 0.6250 & 0.3075 & 9.87 \\
alphaedit L100 (s1) & 0.6300 & 0.2986 & 9.85 \\
alphaedit L100 (s2) & 0.6325 & 0.2919 & 9.87 \\
alphaedit b200 L0 & 0.6275 & 0.2962 & 9.80 \\
alphaedit b200 L1000 & 0.5750 & 0.3372 & 11.90 \\
alphaedit b200 L100 & 0.6250 & 0.3075 & 9.87 \\
alphaedit b50 L0 & 0.6275 & 0.2962 & 9.80 \\
alphaedit b50 L1000 & 0.2275 & 0.3937 & 63.06 \\
alphaedit b50 L100 & 0.6175 & 0.3115 & 9.88 \\
baseline & 0.6275 & 0.2962 & -- \\
damage matched L1000 & 0.5650 & 0.3623 & 12.44 \\
damage response dr0p5 L1000 & 0.6450 & 0.3007 & 10.37 \\
damage response dr0p75 L1000 & 0.6050 & 0.3280 & 10.77 \\
damage response dr1 L1000 & 0.5600 & 0.3421 & 11.90 \\
damage response dr1p5 L1000 & 0.5250 & 0.3655 & 15.68 \\
damage response dr2 L1000 & 0.3725 & 0.4828 & 25.97 \\
lora L0 & 0.6275 & 0.2962 & 9.80 \\
lora L0 (s1) & 0.6275 & 0.2962 & 9.80 \\
lora L1000 & 0.6100 & 0.2662 & 39.23 \\
lora L1000 (s1) & 0.6250 & 0.2717 & 39.88 \\
lora L1000 (s2) & 0.6000 & 0.2494 & 33.17 \\
lora L100 & 0.6650 & 0.2670 & 26.24 \\
lora L100 (s1) & 0.6400 & 0.2733 & 23.17 \\
lora L100 (s2) & 0.6050 & 0.3002 & 23.82 \\
lora gentle L0 & 0.6275 & 0.2962 & 9.80 \\
lora gentle L1000 & 0.6275 & 0.2987 & 28.96 \\
lora gentle L100 & 0.6275 & 0.2932 & 22.07 \\
lora shuffled L0 & 0.6275 & 0.2962 & 9.80 \\
lora shuffled L1000 & 0.3875 & 0.4284 & 237.81 \\
lora shuffled L100 & 0.6225 & 0.2618 & 31.78 \\
memit L0 & 0.6275 & 0.2962 & 9.80 \\
memit L0 (s1) & 0.6275 & 0.2962 & 9.80 \\
memit L1000 & 0.6050 & 0.3106 & 11.49 \\
memit L1000 (s1) & 0.6100 & 0.3160 & 11.30 \\
memit L1000 (s2) & 0.5725 & 0.3401 & 11.11 \\
memit L100 & 0.6225 & 0.3011 & 9.83 \\
memit L100 (s1) & 0.6275 & 0.3034 & 9.82 \\
memit L100 (s2) & 0.6250 & 0.3002 & 9.84 \\
memit shuffled L0 & 0.6275 & 0.2962 & 9.80 \\
memit shuffled L1000 & 0.2825 & 0.5327 & 14.65 \\
memit shuffled L100 & 0.6175 & 0.3097 & 9.83 \\
no edit L0 & 0.6275 & 0.2962 & 9.80 \\
no edit L1000 & 0.6275 & 0.2962 & 9.80 \\
no edit L100 & 0.6275 & 0.2962 & 9.80 \\
random perturb L0 & 0.6275 & 0.2962 & 9.80 \\
random perturb L0 (s1) & 0.6275 & 0.2962 & 9.80 \\
random perturb L1000 & 0.6250 & 0.3129 & 9.90 \\
random perturb L1000 (s1) & 0.6125 & 0.3130 & 10.00 \\
random perturb L1000 (s2) & 0.6325 & 0.2860 & 10.00 \\
random perturb L100 & 0.6225 & 0.3052 & 9.80 \\
random perturb L100 (s1) & 0.6200 & 0.3044 & 9.83 \\
random perturb L100 (s2) & 0.6300 & 0.2927 & 9.83 \\
\bottomrule
\end{tabular}
\caption{Capability measurements at every Qwen checkpoint. The no-edit arm reproduces the baseline exactly at all loads and seeds, which is the check that the pipeline cannot manufacture drift. Three arms reach chance MMLU (0.25).}
\label{tab:degradation}
\end{table}

\bibliographystyle{plainnat}

\end{document}